\documentclass[draftclasofoot, onecolumn]{IEEEtran}
\IEEEoverridecommandlockouts
\usepackage{array}
\usepackage[caption=false]{subfig}
\usepackage[noadjust]{cite}
\usepackage[bookmarks=true,breaklinks=true,colorlinks,linkcolor=red,citecolor=blue,urlcolor=BlueViolet]{hyperref}
\usepackage{multirow}
\usepackage{amsmath,amssymb,amsfonts}

\usepackage{graphicx}
\usepackage{threeparttable}
\usepackage{textcomp}
\usepackage{adjustbox} 
\usepackage{xcolor}
\def\BibTeX{{\rm B\kern-.05em{\sc i\kern-.025em b}\kern-.08em
    T\kern-.1667em\lower.7ex\hbox{E}\kern-.125emX}}
\usepackage{booktabs}
\usepackage{tablefootnote}
\usepackage{float}
\usepackage{algorithmic}
\usepackage{textcomp}
\usepackage{booktabs}

\begin{document}

\title{Neural Canonical Polyadic Factorization for Traffic Analysis}

\author{Wenyu Luo\textsuperscript{*} \quad Yikai Hou\textsuperscript{*} \quad Peng Tang\textsuperscript{*} \quad \thanks{\textsuperscript{*}College of Computer and Information Science, Southwest University, Chongqing, China (1242260715@qq.com, yikaih@email.swu.edu.cn, tangpengcn@swu.edu.cn). This paper is a preprint of a paper submitted to 9th International Conference on Electronic Information Technology and Computer Engineering (EITCE 2025) and is subject to Institution of Engineering and Technology Copyright. If accepted, the copy of record will be available at IET Digital Library.}
}

\maketitle

\begin{abstract}
Modern intelligent transportation systems rely on accurate spatiotemporal traffic analysis to optimize urban mobility and infrastructure resilience. However, pervasive missing data caused by sensor failures and heterogeneous sensing gaps fundamentally hinders reliable traffic modeling. This paper proposes a Neural Canonical Polyadic Factorization (NCPF) model that synergizes low-rank tensor algebra with deep representation learning for robust traffic data imputation. The model innovatively embeds CP decomposition into neural architecture through learnable embedding projections, where sparse traffic tensors are encoded into dense latent factors across road segments, time intervals, and mobility metrics. A hierarchical feature fusion mechanism employs Hadamard products to explicitly model multilinear interactions, while stacked multilayer perceptron layers nonlinearly refine these representations to capture complex spatiotemporal couplings. Extensive evaluations on six urban traffic datasets demonstrate NCPF's superiority over six state-of-the-art baselines. By unifying CP decomposition's interpretable factor analysis with neural network's nonlinear expressive power, NCPF provides a principled yet flexible approaches for high-dimensional traffic data imputation, offering critical support for next-generation transportation digital twins and adaptive traffic control systems.()
\end{abstract}

\begin{IEEEkeywords}
TRAFFIC DATA IMPUTATION, TENSOR DECOMPOSITION, NEURAL TENSOR FACTORIZATION, CANONICAL POLYADIC, LATENT FACTORIZATION OF TENSOR.
\end{IEEEkeywords}

\section{Introduction}

The convergence of 5G-enabled vehicular networks and ubiquitous edge computing infrastructure has catalyzed the generation of petabyte-scale traffic data streams across smart cities \cite{ZhangX24,MiJ23,XuX23,LinM25}. These multimodal observations manifest as irregular hypergraphs combining spatial road segments, temporal slices, mobility metrics (e.g., trajectory density, velocity fields), and heterogeneous sensing sources (loop detectors, drone surveillance, crowdsourced GPS). However, persistent data integrity challenges stem from multi-source asynchrony, transient hardware malfunctions in extreme weather, and privacy-preserving data obfuscation mechanisms, creating compound missing patterns that undermine the reliability of real-time traffic simulation platforms \cite{GC25,BZ24,ChengS24,ChenH25,Lu25,ChenH24,LinM25_,YangH25}. Robust imputation of these multimodal gaps has become pivotal for enabling adaptive traffic flow optimization, connected vehicle coordination, and resilient urban air mobility systems.

Existing approaches typically integrate tensor decomposition \cite{LuoX21_,LuoX23_,TangP24,ChenM24,PengZ22,ChenM25,BiF22,WuD23,WuH22+,LuoX21__,ChenD21,LiW23,LiW22+,ChenM21+,ZhongY23,LuoX21____,WangQ22+,HuL21+,SuX21,YanJ23,QiY21,HuL21,HeY21,HeQ19,SongY22,ZhouY21,LiuZ20} with domain-specific priors (e.g., spatiotemporal smoothness, road network topology constraints) to design completion models, enabling simultaneous capture of global low-rank structures and local spatiotemporal dependencies. Compared to traditional matrix completion methods \cite{LiuZ21, ChenJ21, LuoX19_,LuoX14,LuoX19__,LuoX14__,LuoX18_,LiuZ18, TangM16}, HDI tensor completion more effectively leverages multidimensional relationships and incorporates advanced techniques like graph neural networks and attention mechanisms to model complex traffic patterns. These advancements not only mitigate data sparsity but also provide high-fidelity data for dynamic traffic state estimation and system-wide optimization, playing a pivotal role in enabling fine-grained management of modern transportation infrastructures.

Tensor factorization methods, Canonical Polyadic (CP) decomposition \cite{TG09}, in particular, has emerged as a prominent technique for its capacity to capture multilinear structures in traffic tensor representations. By decomposing data into a sum of rank-one component tensors formed through factor vector outer products, this approach enables efficient dimensionality reduction while explicitly modeling latent interactions across modalities. Each factor vector corresponds to interpretable patterns in specific dimensions, facilitating the identification of dominant traffic flow components and their cross-modal couplings. Yet, conventional CPD models rely on static linear projections, rendering them inadequate for capturing time-varying dependencies or contextual anomalies (e.g., accidents, weather disruptions). Meanwhile, modern neural architectures like transformers \cite{A17} and temporal convolutional networks (TCNs) \cite{ZhenY22} excel at modeling sequential dynamics but often disregard the inherent low-rank structures and interpretable factorizations crucial for traffic system analysis. Recent attempts to merge tensor algebra with deep learning remain limited in adaptively reconciling global tensor constraints with local neural feature learning—a gap that undermines both predictive accuracy and operational interpretability.

Traditional latent factorization of tensor (LFT) approaches \cite{WuD25,WuD21,WuH22,WuH21_,WuH21__,WuH24,TangP24_,LiM21+,WuD22,LuoX21,LuoX16,LuoX18,LuoX22_,ChenJ23,LuoX21+,LuoX22+,LiJ24,ZhongY24,WangQ22,YuanY24,LiW22,LuoX21++,WuH22_,ShiX20,YuanY20,WuD21+,WuD21++,WangJ24+,WuD23+,YuanY24+,YuanY23+,WuH21++,ChenM24++,WuH24+,ChenH24_,XuX25} predominantly rely on multi-linear algebraic frameworks to model spatiotemporal interactions. Linear decomposition methods, such as CP-based nonnegative factorization \cite{TangM16} and its temporal dynamics-aware variants \cite{LuoX20}, demonstrate efficiency in capturing dynamical multi-way correlations. To address data sparsity, density-oriented regularization nonnegative updates \cite{LuoX22__} and alternating direction method-based sequential learning \cite{WuH22__}—leverage constrained optimization to enhance robustness. While these methods achieve interpretable factorizations, their linear assumptions fundamentally limit their capacity to model nonlinear spatiotemporal dependencies (e.g., congestion propagation with threshold effects). 

To address these issues, we propose a novel model named Neural Canonical Polyadic Factorization (NCPF) for traffic data imputation task. Our proposed model aims to make the following contributions: 1) Embedding CP decomposition into a learnable projection mechanism of neural networks to achieve multi-dimensional dense representation learning of sparse traffic data; 2) Combining Hadamard multilinear interaction and stacked neural layer nonlinear optimization to accurately decouple complex spatiotemporal coupling effects. Extensive experimental evaluations demonstrate that NCPF significantly outperforms existing cutting-edge approaches.

\begin{figure*}[t]
\centering
\includegraphics[width=19cm]{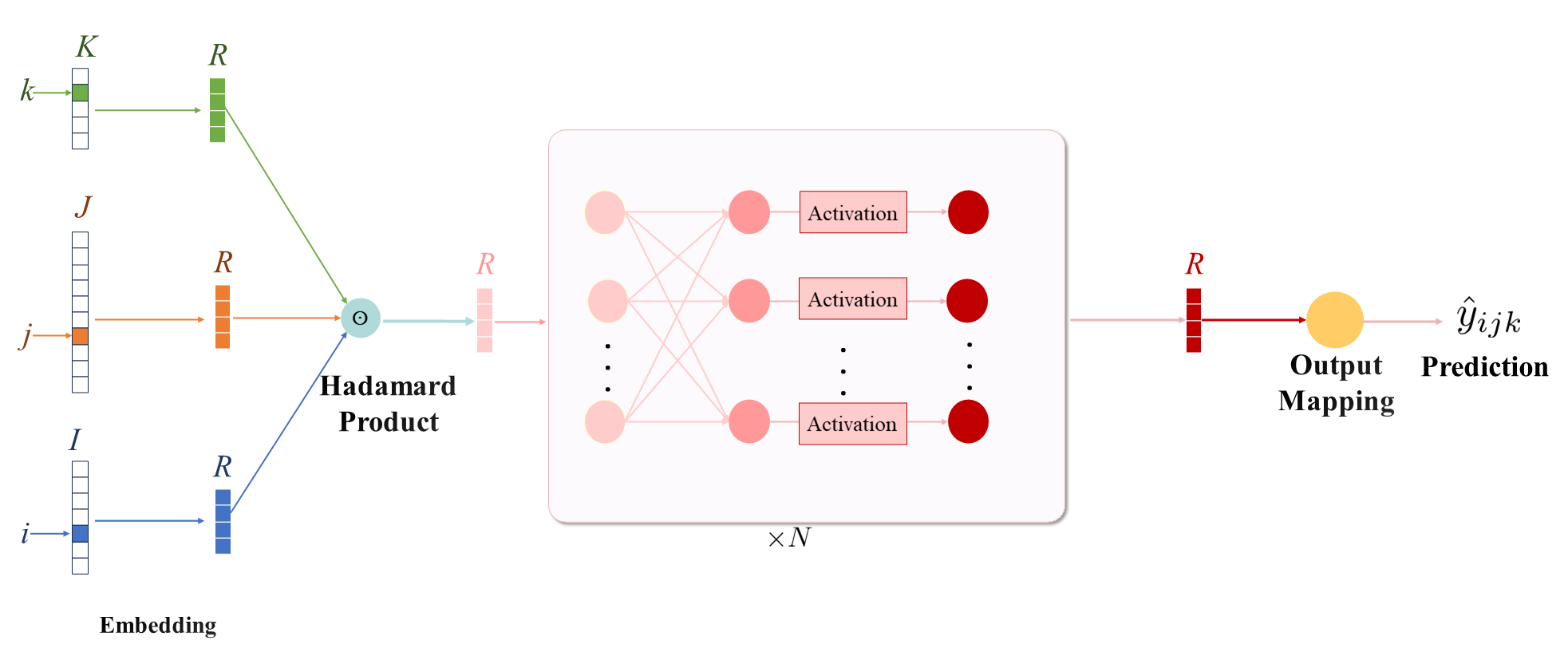}
	\caption{The Neural Canonical Polyadic Factorization (NCPF).}
	\label{fig:model}
\end{figure*}

\section{Methodology }
\subsection{Model Design}

The model architecture diagram is shown in Fig. \ref{fig:model}. The proposed NCPF is engineered for processing multidimensional spatiotemporal tensor data, where input tensor can be represented as $\mathcal{A} \in \mathbb{R}^{M_1 \times M_2\times  \dots \times M_n}$, with each dimension corresponding to a spatial or temporal axis in the coordinate system. To ensure methodological clarity and align with industry-standard practices, the technical exposition will predominantly employ the widely adopted three-order tensor framework. For a three-dimensional tensor $\mathcal{X} \in \mathbb{R}^{I \times J\times   K}$ , the model deals with each tensor element $(i,j,k)$.

The initial step involves converting each component of the triples into high-dimensional binary sparse representations through one-hot encoding. Subsequently, these sparse vectors are projected into lower-dimensional dense embeddings using an embedding layer. So we get three mode embeddings $\mathbf{a}_i$, $\mathbf{b}_j$ and $\mathbf{c}_k$. This process can be illustrated as

\begin{equation}
\mathbf{a}_i = f_{OH}(i)\mathbf{E}_i^\top 
\in \mathbb{R}^R, \mathbf{E}_i \in \mathbb{R}^{I\times J}.
\end{equation}
$f_{OH}(\cdot)$ represents a one-hot operation and $\mathbf{E}_i$ represents a learnable embedding matrix. Then, to simulate the joint effect of traffic flow, speed, and geographic location at a known timestamp, we apply the Hadamard product to perform element-wise multiplication of their feature vectors. Specifically, the fused features are calculated as follows

\begin{equation}
    \mathbf{t}_{ijk} = \mathbf{a}_i \odot \mathbf{b}_j \odot \mathbf{c}_k,
\end{equation}
where $\odot$ denotes Hadamard product. This operation amplifies local interactions while maintaining computational efficiency. Afterwards, the fused input features  $\mathbf{t}_{ijk}$ are linearly mapped and nonlinearly transformed through stacked neuron layers to abstract the spatiotemporal hidden patterns layer by layer, thereby capturing the nonlinear relationship of multi-dimensional coupling in the traffic system. They can be expressed as

\begin{equation}
\mathbf{h}^{(l)} = \alpha(\mathbf{W}^{(l)}\mathbf{h}^{(l-1)}+\mathbf{b}^{(l)}). l=1,2,\dots,L
\end{equation}

$\mathbf{W} \in \mathbb{R} ^{R \times R}$and $\mathbf{b} \in \mathbb{R} ^{R}$ is the weight matrix and bias of each linear layer and is set equal to $\mathbf{t}_{ijk}$. $ \alpha $ represents the activation function such as Sigmoid, RELU, Tanh and so on. However, each activation function has different adaptability to the data set. In the subsequent experimental results, we will discuss the performance of the activation function on different data sets.

Finally, $\mathbf{h}^{(L)}$ will be projected into a single value by a linear transformation. This paper chooses sigmoid activation as output mapping. They can be expressed as

\begin{equation}
    y_{ijk} = \sigma(\mathbf{W}\mathbf{h}^{(L)}),
\end{equation}

where $\sigma$ represents the sigmoid function and $\mathbf{W} \in \mathbb{R} ^{1 \times R}$ denotes the weight matrix of the linear layer. The current approach utilizes a single linear transformation layer to generate the final output, with the main purpose of validating the effectiveness of the neural CP decomposition base model. Future research directions may include replacing this basic projection with more sophisticated alternatives such as layered neural networks, parallelized attention modules, or reconstruction-oriented encoder-decoder topologies. These architectural enhancements could facilitate a comprehensive exploration of the latent tensor structure, enabling deeper investigation of complex spatiotemporal interactions, while exploring the potential complementarity between decomposition techniques and deep representation hierarchies in multidimensional data analysis.

\subsection{Learning Schem}
 To assess the model's efficacy in traffic data completion tasks, we formulate a loss function grounded in the L2 norm between observed traffic measurements $y$ and completed outputs $\hat{y}$. Aligned with the data-driven paradigm that utilizes available observations in set $\Lambda$, this objective function is mathematically expressed as:
 
\begin{align}
\mathcal{E}&=\frac{1}{2}\sum_{y_{ijk} \in \Lambda}(y_{ijk}-\hat{y}_{ijk})^2 \nonumber\\
&=\frac{1}{2}\sum_{y_{ijk} \in \Lambda}\Biggl(y_{ijk}-  \sigma (\mathbf{W}\mathbf{e} _{ijk}^{(n)})\Biggr)^2.
\end{align}
The model parameters are optimized using stochastic gradient descent (SGD) or its enhanced counterparts—including AdaGrad, Adam and RMSProp—within a gradient-driven optimization paradigm. Our implementation adopts the Adam algorithm to attain accelerated convergence rates compared to standard SGD, while preserving prediction fidelity under data sparsity conditions. Empirical results demonstrate that the Adam optimizer achieves a 45.78\% decrease in root mean square error (RMSE) compared to SGD baseline, and requires 84.81\% fewer training iterations to reach convergence thresholds.

\begingroup
\renewcommand\arraystretch{1}
\newcolumntype{D}[1]{>{\centering\arraybackslash}m{#1}}
\newcolumntype{C}[1]{>{\centering\arraybackslash}p{#1}}
\begin{table}[]
\renewcommand\arraystretch{1}
\scriptsize
\centering
\caption{Details of Experiment Datasets}
\label{tab:datas}
\renewcommand\arraystretch{2}
\begin{tabular}{C{2cm}C{2cm}C{2cm}C{2.0cm}C{2cm}C{2cm}}
\noalign{\hrule height 1pt}

\textbf{No.} & \textbf{Dataset} & \textbf{Dimension} & \textbf{Known Count} & \textbf{Density (\%)} \\
\hline
  \textbf{D1} &          Seattle City Speed       & 323×28×288 & 260,467 & 9.9998         \\
  \textbf{D2}  & Hangzhou Flow           & 80×105×28   & 20,976 & 8.9183     \\ 
\textbf{D3}   & New York City Flow           & 30×30×1064 & 97,446 & 7.3957        \\
\textbf{D4}                    & Guangzhou Speed         & 214×61×144  & 185,559  & 9.8713        \\
 \textbf{D5}                    & METR-LA Speed         & 288×119×207  & 651,900  & 9.1891        \\
 \textbf{D6}                    & PEMS-BAY Speed         & 288×181×325  & 1,693,717  & 9.9973        \\
\noalign{\hrule height 1pt}
\end{tabular}
\end{table}
\endgroup

\begin{table}[]
\centering
\caption{THE SUMMARY OF RESULTS}
\label{tab:result}
\renewcommand{\arraystretch}{2}
\footnotesize
\setlength{\tabcolsep}{3.5pt} 
\begin{adjustbox}{max width=\textwidth} 
\begin{tabular}{l *{6}{ccc} c}
\toprule[1.5pt]
\multirow{2}{*}{} & 
\multicolumn{3}{c}{\textbf{D1}} & 
\multicolumn{3}{c}{\textbf{D2}} & 
\multicolumn{3}{c}{\textbf{D3}} & 
\multicolumn{3}{c}{\textbf{D4}} & 
\multicolumn{3}{c}{\textbf{D5}} & 
\multicolumn{3}{c}{\textbf{D6}} \\
\cmidrule(lr){2-4} \cmidrule(lr){5-7} \cmidrule(lr){8-10} \cmidrule(lr){11-13} \cmidrule(lr){14-16} \cmidrule(lr){17-19}
 & \textbf{MAE} & \textbf{MRE} & \textbf{RMSE} 
 & \textbf{MAE} & \textbf{MRE} & \textbf{RMSE} 
 & \textbf{MAE} & \textbf{MRE} & \textbf{RMSE} 
 & \textbf{MAE} & \textbf{MRE} & \textbf{RMSE} 
 & \textbf{MAE} & \textbf{MRE} & \textbf{RMSE}
 & \textbf{MAE} & \textbf{MRE} & \textbf{RMSE} \\
\midrule[0.8pt]
\textbf{M1} 
& 7.3406 & 0.2282 & 10.5573 
& 70.7900 & 1.9060 & 137.2378 
& 11.6017 & 4.2979 & 14.9034 
& 6.5716 & 0.2198 & 8.6559 
& 5.0974 & 0.1181 & 8.2495
& 8.9371  & 0.2377 & 12.1656 \\

\textbf{M2} 
& 5.4252 & 0.1356 & 7.8008 
& 25.2415 & 0.2217 & 59.7254 
& 5.0889 & 0.6151 & 11.0201 
& 3.6128 & 0.1238 & 5.1250 
& 3.9895 & 0.0816 & 6.1690
& 6.0145 & 0.1482 & 8.8401 \\

\textbf{M3} 
& 4.9711 & 0.1293 & 7.2974 
& 24.0271 & 0.2295 & 53.3764
& 5.0559 & 0.6146 & 10.9956 
& 3.6676 & 0.1256 & 5.1623 
& 3.3521 & 0.0702 & 5.5257
& 5.7189 & 0.1426 & 8.6033 \\

\textbf{M4} 
& 5.4649 & 0.1371 & 7.8439 
& 24.6459 & 0.2235 & 58.7152 
& 5.1034 & 0.6310 & 10.9229 
& 3.6610 & 0.1257 & 5.1598 
& 3.9887 & 0.0821 & 6.1943
& 5.9879 & 0.1476 & 8.8081 \\

\textbf{M5} 
& 4.2169 & 0.1064 & 6.5043 
& \textbf{23.4148} & 0.2194 & \textbf{51.0982} 
& 4.6252 & 0.5699 & 9.7396 
& 3.2624 & \textbf{0.1123} & \textbf{4.7383} 
& 2.9225 & 0.0619 & 5.0976
& 4.9328 & 0.1228 & 7.8116 \\

\textbf{M6} 
& 4.5040 & 0.1148 & 6.7490
& 24.9185 & 0.2175 & 55.8657 
& \textbf{4.2742} & \textbf{0.5307} & \textbf{8.4263 }
& 3.2746 & 0.1129 & 4.7575 
& 3.1842 & 0.0667 & 5.3216
& 4.7763 & 0.1184 & 7.5979 \\

\textbf{M7} 
& \textbf{4.0574} & \textbf{0.1026} & \textbf{6.3051} 
& 24.7449 & \textbf{0.2166} & 55.2637 
& 4.3805 & 0.5389 & 8.6662
& \textbf{3.2493} & 0.1126 & 4.7398
& \textbf{2.7973} & \textbf{1.0127} & \textbf{4.8448}
& \textbf{4.6254} & \textbf{0.1167} & \textbf{7.5346} \\
\bottomrule[1.5pt]
\end{tabular}
\end{adjustbox}
\end{table}

\section{Experiments}
\subsection{Experiment Setting}
To verify the performance of the model, this study conducted experiments based on six real urban traffic data sets (see Tab. \ref{tab:datas} for detailed information on the data sets). In order to solve the problem that the original data distribution is skewed and does not conform to the probability assumption of the low-rank decomposition model, logarithmic transformation and minimum-maximum normalization are used to preprocess the data: first, the high-value interval is compressed by logarithmic transformation to alleviate the right-skewed distribution characteristics and make the data distribution closer to the normal distribution; then the minimum-maximum normalization is applied to linearly map the data to the [0,1] interval to eliminate the interference of dimensional differences on model training. The above method effectively corrects the non-Gaussian characteristics of dynamic data such as traffic flow and speed, enhances its compatibility with the low-rank matrix decomposition assumption, and provides a standardized data basis for cross-dataset performance verification. 

The NCPF framework is comprehensively benchmarked against state-of-the-art methods to validate its effectiveness in imputing missing traffic data. The baselines are listed as follows: 

1) M1: Lightweight Graph Convolutional Network (Light-GCN) \cite{HeX20};

2) M2: a Nonnegative Canonical Polyadic-based Latent Fac-torization of Tensor (NNCP) model \cite{ZhangW14};

3) M3: a Canonical Polyadic-based Latent Factorization of Tensor model with Cauchy loss-based objective function(CTF) \cite{YeF21};

4) M4: Biased Nonnegative Latent Factorization of Tensor(BNLFT) model with SLF-NMUT and ADM-STL Opti-mization \cite{LuoX20};

5) M5: Bayesian Tensor Time Series Analyzer with VAR-Gibbs Integration (BTTF)\cite{ChenX22};

6) M6: Neural Collaborative Filtering \cite{HeX17};

7) M7: our Neural Canonical Polyadic Factorization (NCPF).

To ensure equitable benchmarking, the latent dimension sizes for neural network-based approaches (M1, M6) and the tensor ranks for LFT-based methods (M2-M5) are uniformly configured to 5. Furthermore, the LFT framework variants (M2-M4) incorporate performance-enhancing techniques through algorithmic integration of SLF-NMUT and ADM-STL modules. Notably, due to the architectural absence of built-in constraints like sigmoid normalization in neural architectures, the preliminary predictions of M2-M4 risk exceeding the normalized [0,1] operational range. To maintain analytical parity, we implement rigorous post-processing clipping on these models' outputs, enforcing strict value confinement within the target interval before inverse normalization. Hyperparameter optimization employs exhaustive grid search across all configuration spaces to ensure peak performance. Addressing stochastic initialization effects, each experimental condition is replicated across ten randomized trials, with final metrics calculated as arithmetic means of all executions.

The experiments were performed on a platform with a 2.50-GHz 13th Gen Intel(R) Core(TM) i5-13400F CPU and one NVIDIA GeForce RTX3050 GPU with 32-GB RAM. All the model are implemented with Python 3.10.12 and Pytorch 2.4.1. This paper employs three evaluation metrics: mean absolute error (MAE), mean relative error (MRE), and root mean square error (RMSE).

\begin{figure}[t]
\centering
\subfloat[]
{\label{fig:resulta}
\includegraphics[width=11cm]
{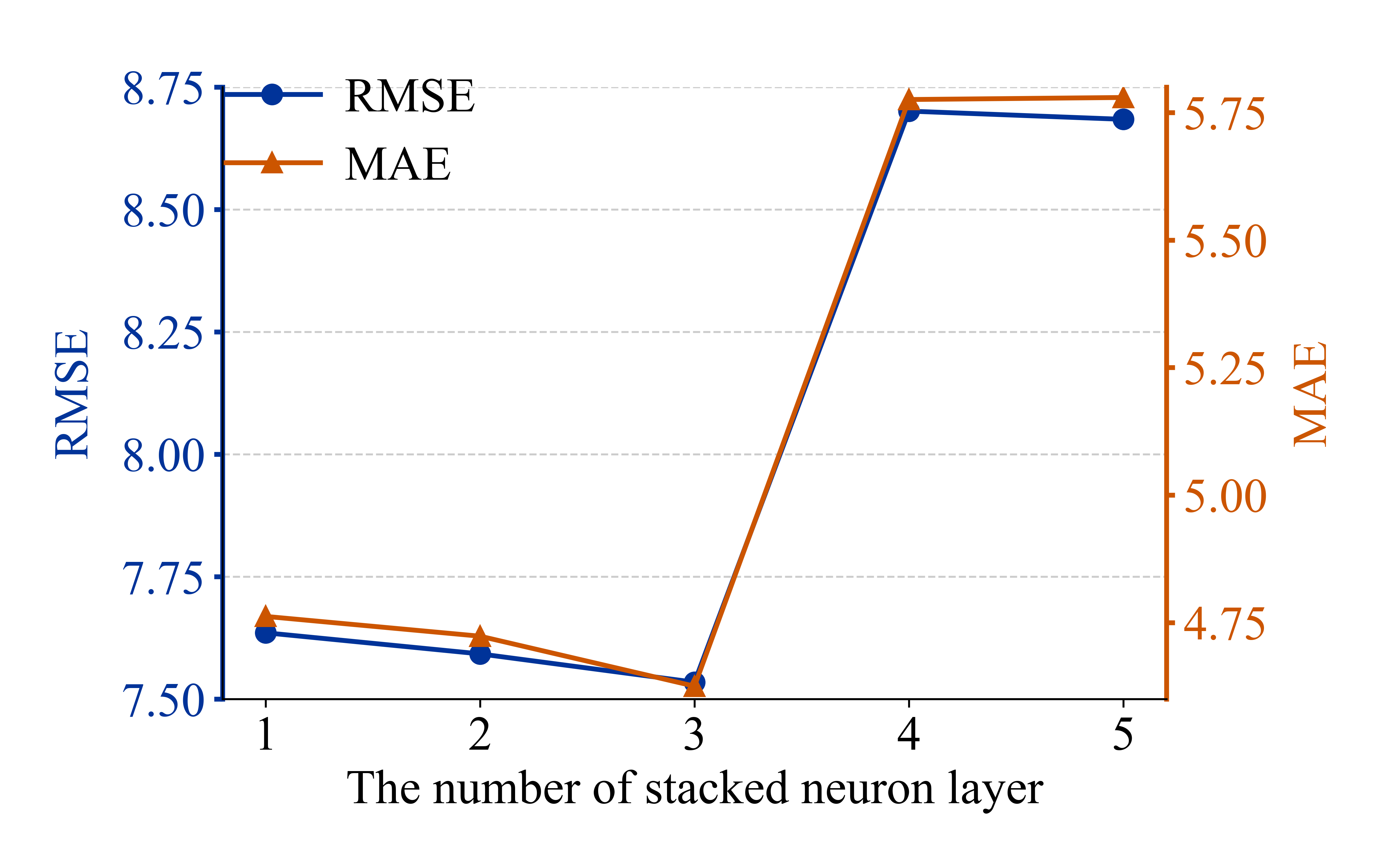} }
\quad
\subfloat[]
{\label{fig:resultb}
\includegraphics[width=11cm]
{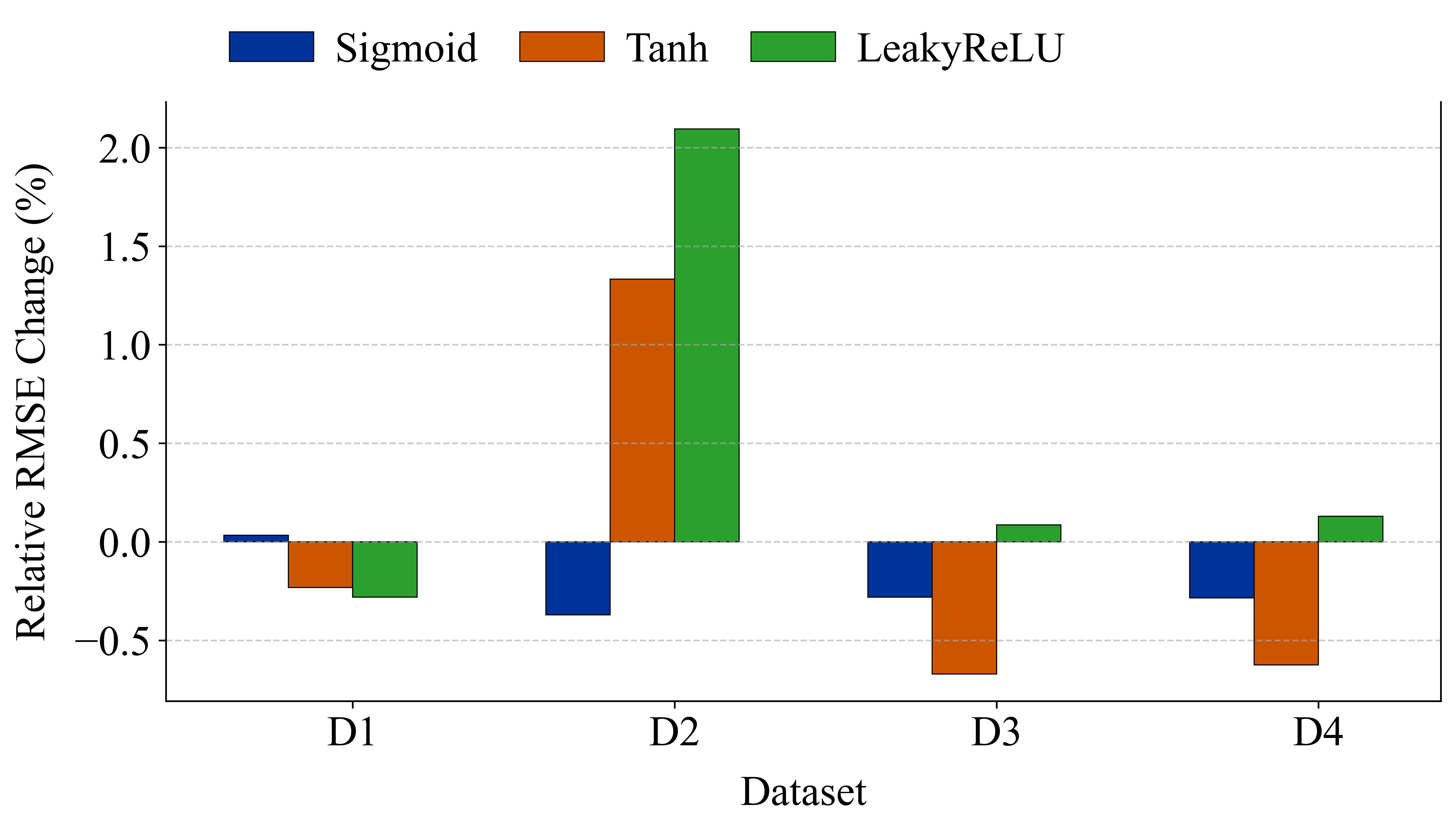}} \\
\caption{The effect of the number of stacked neuron layer and different activation function of NCPF.
}
\label{fig:result}
\end{figure}

\subsection{Result Analysis}

1) \textit{Combining CP decomposition with neural network architecture and introducing nonlinear transformation can effectively complete traffic spatiotemporal data}\textit{: }The experimental results are shown in Tabble. \ref{tab:result}, indicating that the proposed neural canonical multivariate factorization (NCPF) model exhibits better completion performance than existing can effectively capture the complex spatiotemporal coupling relationship in traffic data, especially in scenarios with high sensor missing rate, and its hierarchical feature fusion mechanism significantly improves the reconstruction accuracy. Across six benchmark datasets (D1-D6), the proposed model sets new performance benchmarks, outperforming traditional linear prediction methods (LFT) (M2-M5) and neural network-based models (M1, M6) on multiple datasets. Particularly noteworthy is its performance on the larger PEMS-BAY speed dataset (D6), where our solution achieves significant RMSE reductions of 41.27\%, 21.46\%, 12.32\%, 21.79\%, 4.96\%, and 8.96\%, over M1-M6, respectively. The modeling advantage of the proposed framework is further demonstrated on the small dataset D1, where it achieves a 3.86\% higher MAE than the highest baseline M5 through an effective ensemble CP decomposition architecture and nonlinear transformations.

2) \textit{The number of layer of the multilayer perceptron has a non-monotonic relationship on the completion effect}\textit{:} The experimental results show that the depth of the multilayer perceptron and the missing value filling performance show a significant non-monotonic correlation: as shown in Fig. \ref{fig:resulta}, when the number of network layers gradually increases from 1 to 3 layers, the model effectively captures the spatiotemporal dependencies in the traffic state through hierarchical feature abstraction, and the RMSE and MAE indicators are optimized from 7.63/4.76 to 7.53/4.62, respectively. This improvement stems from the ability of moderate depth to collaboratively model the nonlinear interactions between road network topological constraints and time-varying traffic patterns; however, when the number of layers exceeds 3, the model performance collapses in the opposite direction, and the RMSE and MAE increase by 5.04\% and 6.71\% respectively in the 5-layer structure. The essence of this can be attributed to the information attenuation effect caused by multi-layer linear transformations during gradient propagation, which makes it difficult for the underlying spatiotemporal features to be effectively updated. The excessive memory of sparse observation noise in the deep parameter space leads to a distorted representation of the potential traffic pattern. At the same time, an overly complex architecture destroys the  baseline methods on multi-city traffic datasets. By integrating the implicit factor analysis of tensor decomposition with the nonlinear expression ability of neural networks, the model generalization boundary of the model. This U-shaped curve that first decreases and then increases essentially reflects the game equilibrium between the complexity of traffic system dynamics and the capacity of deep learning models. The three-layer structure can balance the algebraic interpretability of CP decomposition and the nonlinear expression ability of neural networks. The probability of deep networks falling into local extreme values increases due to parameter redundancy, which verifies the engineering effectiveness of the "moderate depth" principle in the task of intelligent traffic  data filling.

3) \textit{Different activation functions perform differently on various datasets}\textit{:} According to the characteristic differences (dimension, time span, density) of the six traffic data sets in Table \ref{tab:result}, the performance of different activation functions shows significant differentiation. Because the effect of the ReLU activation function is the least obvious and there is no most suitable data set, we use the result processed by the ReLU function as the baseline, and then calculate the relative percentage change of the data of other activation functions. The process can be denoted as

\begin{equation}
  \Delta \% _{act,dataset}=  \frac{M_{act, dataset}-M_{baseline,dataset}}{M_{baseline, dataset}}\times100\%
\end{equation} 
 where $act$ denotes different activation function. The results are shown in Fig. \ref{fig:resultb}: LeakyReLU is most suitable for D1, Sigmoid is most suitable for D2, and Tanh is most suitable for D3 and D4. This phenomenon is essentially the result of dynamic adaptation of data characteristics (density/time series/space) and the nonlinear response mode of the activation function (sparseness/smoothness/gradient stability). These phenomena also verify the necessity of dynamically selecting the activation function according to the data distribution (sparseness/symmetry) and the task goal (regrsion/probability /prediction).

\section{Conclusions and future works}\label{con}

This paper proposes the Neural Canonical Polynomial Factorization (NCPF) framework, a novel framework that synergizes tensor factorization with deep neural networks to solve the problem of multimodal spatiotemporal traffic data imputation. By embedding the Canonical Polynomial Factorization (CP) into a multi-layer neural architecture, NCPF reinterprets traditional tensor algebra operations as neural components, enabling dynamic learning of latent correlations while preserving the inherent multilinear structural priors of traffic data. The model establishes a hierarchical network space where shallow layers capture spatiotemporal fusion representations through tensor factor embedding and hardmard products, while deep layers refine these representations through nonlinear transformations to model complex interactions. This mechanism enables the model to go beyond the linear subspace assumption of traditional CP factorization while avoiding the opacity of purely data-driven deep learning methods.

Future research directions focus on enhancing the scalability and temporal modeling capabilities of the framework. First, extending NCPF to the domain of high-order tensor factorization will address emerging urban perception modalities such as multi-UAV traffic monitoring networks and high-dimensional air traffic patterns. Second, the integrated time series modeling architecture (including LSTM and GRU networks) can explicitly capture the long-range temporal dependencies in traffic flow evolution, especially for abrupt phase changes caused by accidents or weather events. This dual-path extension will further combine the structural advantages of tensor algebra with the dynamic modeling capabilities of deep learning to ad vance the development of robust intelligent transportation systems.

\end{document}